\documentclass{article}

\PassOptionsToPackage{numbers, compress}{natbib}

    \usepackage[preprint]{neurips_2023}

\usepackage[utf8]{inputenc} 
\usepackage[T1]{fontenc}    
\usepackage{hyperref}       
\usepackage{url}            
\usepackage{booktabs}       
\usepackage{amsfonts}       
\usepackage{nicefrac}       
\usepackage{microtype}      
\usepackage{xcolor}         
\usepackage{graphicx}
\usepackage{amsmath}

\usepackage{amssymb}
\usepackage{multirow}
\title{EnSiam: Self-Supervised Learning With \\ Ensemble Representations}

\author{
  Kyoungmin Han \quad Minsik Lee \\
  Department of Electrical and Electronic Engineering\\
  Hanyang University ERICA \\
  Ansan 15588, Korea \\
  \texttt{\{gkssrudalls,mleepaper\}@hanyang.ac.kr} \\
}

\begin{document}

\maketitle

\begin{abstract}

Recently, contrastive self-supervised learning, where the proximity of representations is determined based on the identities of samples, has made remarkable progress in unsupervised representation learning.
SimSiam is a well-known example in this area, known for its simplicity yet powerful performance. However, it is known to be sensitive to changes in training configurations, such as hyperparameters and augmentation settings, due to its structural characteristics.
To address this issue, we focus on the similarity between contrastive learning and the teacher-student framework in knowledge distillation. Inspired by the ensemble-based knowledge distillation approach, the proposed method, EnSiam, aims to improve the contrastive learning procedure using ensemble representations.
This can provide stable pseudo labels, providing better performance. Experiments demonstrate that EnSiam outperforms previous state-of-the-art methods in most cases, including the experiments on ImageNet, which shows that EnSiam is capable of learning high-quality representations.

\end{abstract}

\section{Introduction}

Self-supervised learning (SSL) has shown notable achievements in recent years, and its performance approaches that of supervised learning \cite{wu2018unsupervised, zhuang2019local, chen2020simple, gidaris2018unsupervised, hjelm2018learning, yang2016joint, zheng2021ressl}. SSL can utilize large amounts of data without human annotations, and it is very promising in situations where labeling costs are prohibitively expensive or obtaining data-label pairs is difficult due to the nature of the problem. The well-trained representation produced by SSL can be utilized in transfer learning to enhance performance in downstream tasks.

In the early days of SSL, unsupervised representation learning methods based on predefined pretext tasks were studied \cite{doersch2015unsupervised, noroozi2016unsupervised, zhang2016colorful, feng2019self, oord2018representation}. Afterward, contrastive learning became the mainstream of SSL \cite{chen2020simple, chen2020big, he2020momentum, chen2020improved, kalantidis2020hard, hu2021adco}, where augmented samples of the same instance become close to each other in the embedding space while those of different instances become far apart. The negative samples, i.e., samples from different instances, were important in addressing the collapse problem, which indicates a situation where a trained network always produces the same representation. Unfortunately, contrastive learning methods that utilize negative samples tend to depend highly on the quality of negative samples, such as their size or diversity, which is difficult to maintain. Accordingly, SSL methods depending solely on positive samples without any negative samples have emerged to resolve this problem. Well-known examples of this approach include BYOL \cite{grill2020bootstrap} and SimSiam \cite{chen2021exploring}. Especially, SimSiam can learn high-quality representations even though it has a relatively simple structure, thanks to the asymmetric applications of a stop gradient operation and a prediction head. However, SimSiam can exhibit somewhat unstable training behaviors, e.g., it can fail to converge or be sensitive to slight changes in training configurations such as augmentation settings or hyperparameters \cite{bai2022directional}.

In this paper, we propose a new SSL method to address the above sensitivity issue.
We note that the structure of contrastive learning is similar to the teacher-student framework of knowledge distillation \cite{hinton2015distilling}. Asymmetric contrastive SSL methods such as MoCo \cite{he2020momentum, chen2020improved}, BYOL, and SimSiam all have branches that do not involve gradient flow. These branches provide pseudo labels, which is similar to the role of teacher networks in knowledge distillation (KD). The remaining branches are designed to learn from the pseudo labels provided by the `teacher' branches, which serve as targets.\footnote{In particular, self-distillation methods\cite{zhang2019your, kim2021self} share the same network for the teacher and student, which is even similar to the framework of asymmetric contrastive SSL.} 
In this context, we focus on the fact that some KD methods adopt ensemble techniques in the teacher part \cite{fukuda2017efficient, yang2019training, asif2019ensemble} to provide better pseudo labels.
Inspired by this, we propose to use ensemble representations to improve SimSiam. This is accomplished by generating multiple augmented samples from each instance and then utilizing their ensemble as a stable pseudo label.

To confirm that ensemble representations have actual benefits in the training procedure, we also devise a variant of SimSiam that generates more than two augmented samples for each instance. 
This variant is henceforth referred to as SimSiam $K$-aug, where $K$ indicates the number of augmentations for each instance.\footnote{Based on this terminology, the original SimSiam is identical to SimSiam 2-aug.} By comparing the proposed method, EnSiam, with SimSiam $K$-aug, we confirm that the performance gain of EnSiam is more than simply increasing the number of augmentations: First, we analyze the gradients of EnSiam and SimSiam $K$-aug in Section \ref{Grad_of_Ensiam}, demonstrating that the use of ensemble representations can be understood in terms of variance reduction.
We then confirm that this indeed leads to better performance in experiments. Furthermore, we verify the performance of EnSiam under various configurations. In particular, EnSiam can perform more robustly under stronger augmentation settings.
In the experiments, the proposed method achieves the highest performance in linear evaluation on various datasets and transfer learning for object detection, confirming that EnSiam can learn high-quality representations in an unsupervised manner.

\section{Related Works}
\label{Related Work}

\textbf{Predefined pretext task.} SSL for representation learning has been extensively studied based on various pre-defined pretraining tasks. Image patch prediction methods \cite{doersch2015unsupervised, noroozi2016unsupervised, kim2018learning} perform representation learning based on various pre-defined patch-based prediction problems. Colorization \cite{zhang2016colorful, larsson2017colorization} and inpainting \cite{pathak2016context} have also been studied to extract visual context for learning image representations. Additionally, rotation prediction, i.e., predicting random rotations applied to input images, was shown to be effective in image representation learning \cite{gidaris2018unsupervised, feng2019self}.
There have also been studies based on future prediction, where a pretext task is defined to predict the future of a given input data \cite{oord2018representation, henaff2020data, lorre2020temporal}. Pretext-task-based SSL has been studied not only for images but also for diverse data such as videos \cite{wang2015unsupervised, duan2022transrank}, keypoints \cite{jakab2020self}, human skeletons \cite{lin2020ms2l}, and sketches \cite{bhunia2021vectorization}, considering the corresponding data characteristics.

\textbf{Mutual information.}
Several methods have also attempted to learn high-quality representations by maximizing mutual information \cite{hjelm2018learning, bachman2019learning}. One example is contrastive multiview coding (CMC) \cite{tian2020contrastive}, which maximizes the mutual information between multiple views. Methods for solving the problem of informational collapse have also been actively studied. For instance, W-MSE \cite{ermolov2021whitening} uses whitening or Karhunen-Loève transform on representations to achieve uniformly distributed representations, while Barlow Twins \cite{zbontar2021barlow} aims to resolve the collapse problem by normalizing the cross-correlation matrix of the representation vectors so that it becomes similar to the identity matrix.

\textbf{Clustering.}
SSL based on clustering in the embedding space has also been actively studied \cite{yang2016joint, asano2019self, caron2019unsupervised, vedaldi2020self, li2020prototypical}. DeepCluster, for example, performs self-supervised learning by providing pseudo labels generated from $k$-means clustering of the previous iteration's representations \cite{caron2018deep}. Furthermore, SwAV proposes an online clustering formulation and achieves balanced cluster assignment through the Sinkhorn-Knopp transform \cite{caron2020unsupervised}. Unfortunately, clustering methods require significant resources to achieve high performance because it requires maintaining either queues, memory banks, or large batch sizes.

\begin{figure}
  \centering
  \includegraphics[width=\columnwidth]{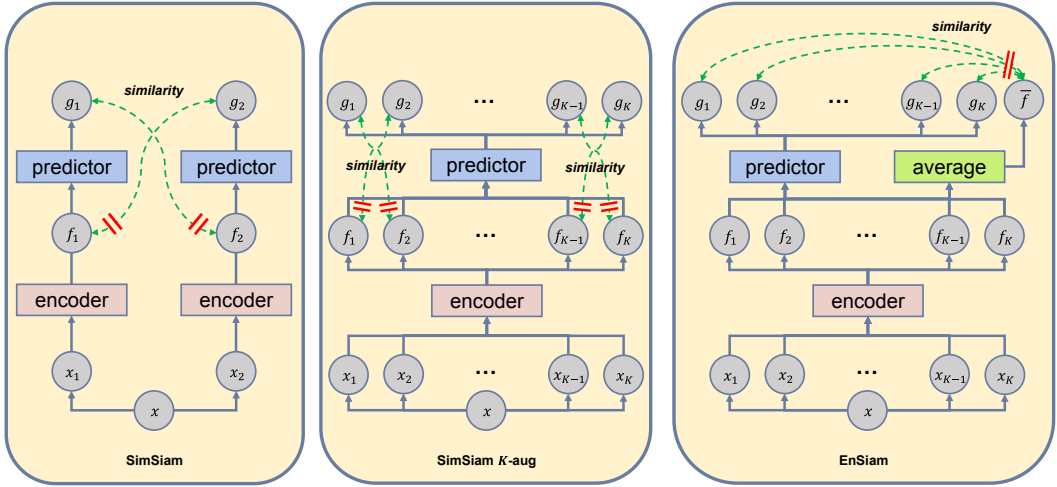}
  \caption{Structure overview. The leftmost is SimSiam, the middle is a simple variant of SimSiam, and the rightmost is the proposed method. All three frameworks use stop gradient operations on the branch providing pseudo labels. In case of EnSiam, the stop gradient operation, represented by a red double line, is applied to $\bar{f}$, which is the average of $f_k$. Both EnSiam and SimSiam $K$-aug generate $K$ samples through augmentation and pass them to the same encoder and predictor. The models are then trained to maximize the similarities described by green dashed lines.}
  \label{systemoverview}
\end{figure}

\textbf{Contrastive learning.}
Contrastive learning has emerged as one of the most prominent approaches in the field of SSL. Fundamentally, contrastive learning is designed based on the assumption that transformed versions of the same sample should have similar representations in the embedding space, while those of different samples should differ more largely.

SimCLR \cite{chen2020simple, chen2020big} is one of the representative methods in contrastive learning. However, it has the drawback of requiring a large number of negative samples. Using a large number of negative samples in one batch increases the computational cost, making it difficult to train in environments with limited resources. Another similar method is MoCo \cite{he2020momentum, chen2020improved}, which employs a memory bank that serves a similar role to the negative samples of SimCLR. While MoCo has a problem of initial feature instability caused by the memory bank, they solve the issue through the moving average update of the encoder network. However, MoCo still requires a large memory bank to learn high-quality representations.

To overcome the issue induced by negative samples, BYOL \cite{grill2020bootstrap} introduced a predictor and a momentum encoder, achieving high-quality representations based only on positive samples. On the other hand, SimSiam \cite{chen2021exploring} achieved a similar goal with a simpler structure consisting of a predictor and a stop gradient operation. VICReg \cite{bardes2021vicreg} introduced a regularizer to alleviate the collapse problem in contrastive learning. Recently, ReSSL \cite{zheng2021ressl} extended the conventional contrastive learning approach to improve performance. Instead of directly comparing the similarity between feature embeddings, it focuses on comparing the similarity of relationships between embeddings within a batch.
A downside of ReSSL is that it has quadratic computational complexity with respect to (w.r.t.) the batch size when generating the relationship distribution.

\textbf{Knowledge distillation.}
Knowledge distillation (KD) has been studied to enhance the training of new models by transferring the knowledge of an existing network that is well-trained \cite{hinton2015distilling,  chen2017learning, yim2017gift, heo2019knowledge}. KD is usually composed of a teacher network and a student network. The teacher network provides some form of pseudo-labels to the student network. There is also a variant to this approach called self-distillation, where a single network becomes both the teacher and student: In \cite{zhang2019your}, the knowledge from the deeper layers of a network is distilled to its shallower layers. Meanwhile,  PS-KD \cite{kim2021self} utilizes progressive self-distillation to gradually soften hard targets during the training process. The multi-teacher methods that utilize the ensemble of multiple teacher networks are studied as well. Fukuda et al. \cite{fukuda2017efficient} employed a random selection process, choosing one teacher network at each iteration. In order to transfer knowledge based on features from multiple teachers, EKD \cite{asif2019ensemble} introduced additional teacher branches to the student networks, mimicking the intermediate features of the teachers.

\section{Proposed Method}
\label{Proposed Method}

In this section, we first revisit SimSiam \cite{chen2020simple}, and introduce the proposed method, EnSiam. We also provide an analysis of the gradient of EnSiam, showing that the proposed method can be understood in terms of variance reduction.

\subsection{SimSiam and Its Simple Variant}

Let $x$ be a sample from an unlabeled dataset $X$. In contrastive learning, multiple augmented views (usually two) are generated from a single sample. $t(\cdot)$ denotes the augmentation transform that is randomly selected from a predefined transform space $T$. Given an input sample $x$, for example, one can generate two augmented views $x_1 \triangleq t_1(x)$ and $x_2 \triangleq t_2(x)$, where $t_1(\cdot)$ and $t_2(\cdot)$ indicate two transforms sampled independently from $T$. Each generated view passes through the encoder function $f(\cdot)$ and the predictor function $g(\cdot)$, i.e., $f_1 \triangleq f(x_1)$, 
 $f_2 \triangleq f(x_2)$,  $g_1 \triangleq g(f_1)$, and $g_2 \triangleq g(f_2)$. The loss function of SimSiam for $x$ is then defined as follows:
\begin{equation}
L_x = \frac{1}{2} \{ D(g_1, \varphi(f_2)) + D(g_2, \varphi(f_1))\}.
\label{eqn:Simsiam_baseline}
\end{equation}
Here, $\varphi(\cdot)$ indicates the stop gradient operation. In our terminology, the branches going through $\varphi(\cdot)$ serve as the teacher branches. The function $D(\cdot, \cdot)$ refers to the distance metric, e.g., cosine distance. The overall loss function becomes $L = \mathbb{E}_x[L_x]$ where $\mathbb{E}_x[\cdot]$ denotes expectation over $x$, where the expectation is approximated by a sample average in practice.

We can express a generalized version of (\ref{eqn:Simsiam_baseline}) considering more than one augmented pair, as follows:
\begin{equation}
L_x = \frac{1}{K} \sum_{k=1}^{K} D(g_k, \varphi(f_{N(k)})).
\label{eqn:Simsiam_reformed}
\end{equation}
Here, the function $N(k) \triangleq 2 \lceil k / 2 \rceil - 1 + (k \mod 2)$ denotes the index of the other sample in similarity computation, e.g., $N(1) = 2$ and $N(2) = 1$. $K$ indicates the number of augmentations generated for a single instance, and it is assumed to be an even number. This variant is called as SimSiam $K$-aug throughout this paper. We can see that it is identical to the original SimSiam loss if $K=2$. The comparison of the original SimSiam and SimSiam $K$-aug is illustrated in Figure \ref{systemoverview}.

Note that both (\ref{eqn:Simsiam_baseline}) and (\ref{eqn:Simsiam_reformed}) are stochastic approximations of the following function:
\begin{equation}
L_x = \mathbb{E}_{t_1, t_2}[ D(g(f(t_1(x))), \varphi(f(t_2(x)))) ] = \mathbb{E}_{t_1, t_2}[ D(g_1, \varphi(f_2)) ],
\label{eqn:Simsiam_true}
\end{equation}
where $t_1$ and $t_2$ simply indicate that they are two independent transforms. The only difference between SimSiam and SimSiam $K$-aug is the number of augmentation pairs sampled to approximate this expectation. Hence, (\ref{eqn:Simsiam_true}) will be used in the analysis of Section \ref{Grad_of_Ensiam}.

\subsection{Ensemble representations for SimSiam}

The main idea of the proposed method is to replace the pseudo labels in SimSiam with the expectations of the projected representations over transform $t$. $\bar{f} \triangleq \mathbb{E}_t[f(t(x))]$ becomes the new pseudo label in the loss function of EnSiam:
\begin{equation}
\bar{L}_x = \mathbb{E}_{t}[ D(g(f(t(x))), \varphi(\mathbb{E}_t[f(t(x))])) ] = \mathbb{E}_{t}[ D(g, \varphi(\bar{f})) ].
\label{eqn:Ensiam_Loss}
\end{equation}
Here, a bar sign is added to $\bar{L}_x$ to separate it from (\ref{eqn:Simsiam_true}). Note that this expectation only involves a single $t$.
In reality, it is intractable to calculate the expectation over $t$. Hence, we approximate $\bar{f}$, as well as $\bar{L}_x$, using sample averages:
\begin{equation}
    \begin{split}
    \bar{f} &\approx \frac{1}{K} \sum_{k=1}^{K} f_k, \\
    \bar{L}_x &\approx \frac{1}{K} \sum_{k=1}^{K} D(g_k, \varphi(\bar{f})).
    \end{split}
\label{eqn:fbar_sampleexp}
\end{equation}

The overview of EnSiam is shown in Figure \ref{systemoverview}. The proposed method can provide stable pseudo labels compared to the original SimSiam. To demonstrate this, we compare SimSiam $K$-aug and EnSiam by analyzing the variances of the gradients in the following section and showing the empirical results in Section \ref{effects_of_hyperparameter}.

\subsection{Analysis of the Variance of Gradients}
\label{Grad_of_Ensiam}

Here, we provide an analysis regarding the variance of gradients. From this analysis, we can confirm that EnSiam has an effect of variance reduction. This can help improve the stability of the training procedure, as confirmed in Section \ref{Empirical Study}.

The loss functions of SimSiam and EnSiam share a similar structure, where only the targets differ. Accordingly, they share similar derivations. Suppose we have a common loss function $D(a, \varphi(b))$ where $a$ and $b$ are two independent random vectors which are also functions of $x$ and parameter $\theta$. Let $\Gamma(a, b)$ be the gradient of this function w.r.t. $\theta$, i.e.,
\begin{equation}
    \Gamma(a, b) \triangleq \nabla_\theta D(a, \varphi(b)) = \left(\frac{\partial D(a, b)}{\partial a} \frac{\partial a}{\partial \theta}\right)^\top.
    \label{eqn:grad1}
\end{equation}
We assume that $\Gamma$ is a smooth function and $b$ is concentrated near $\bar{b} \triangleq \mathbb{E}[b]$, which is plausible since $b$ is essentially the projected feature representation of $x$ transformed by a random $t$. Accordingly, we approximate $\Gamma$ using the first-order Taylor series expansion on $b$ around $\bar{b}$:
\begin{equation}
    \Gamma(a, b) \approx \Gamma(a, \bar{b}) + \frac{\partial \Gamma(a, \bar{b})}{\partial b} (b - \bar{b}).
    \label{eqn:taylor}
\end{equation}
Then, the variance of $\Gamma$ becomes
\begin{equation}
    Var\!\left[\Gamma(a, b)\right] \approx Var\!\left[\Gamma(a, \bar{b})\right] + \mathbb{E}\!\left[\frac{\partial \Gamma(a, \bar{b})}{\partial b} Var[b] \frac{\partial \Gamma(a, \bar{b})}{\partial b}^\top \right].
    \label{eqn:simplified_with_var_cov}
\end{equation}
Here, $Var[\cdot]$ is a covariance matrix since $\Gamma$ and $b$ are vectors. This suggests that the overall variance is decomposed into two terms. Depending on SimSiam or EnSiam, $a$ and $b$ differ.

For EnSiam, $a$ and $b$ are $g$ and $\bar{f}$, respectively. Accordingly, $Var[b]$ in (\ref{eqn:simplified_with_var_cov}) vanishes, and the variance of the gradient becomes simply $Var\!\left[\Gamma(g, \bar{f})\right]$.
On the other hand, those of SimSiam are $g_1$ and $f_2$, respectively, and therefore the variance becomes
\begin{equation}
    \begin{split}
    Var\!\left[\Gamma(g_1, f_2)\right] &\approx Var\!\left[\Gamma(g_1, \bar{f}_2)\right] + \mathbb{E}\!\left[\frac{\partial \Gamma(g_1, \bar{f}_2)}{\partial g_1} Var[f_2] \frac{\partial \Gamma(g_1, \bar{f}_2)}{\partial g_1}^\top \right] \\
    &= Var\!\left[\Gamma(g, \bar{f})\right] + \mathbb{E}\!\left[\frac{\partial \Gamma(g, \bar{f})}{\partial g} Var[f] \frac{\partial \Gamma(g, \bar{f})}{\partial g}^\top \right],
    \end{split}
\label{eqn:alternation_with_ai}
\end{equation}
where $\bar{f}_2 \triangleq \mathbb{E}[f_2] = \bar{f}$ is the mean of $f_2$. Note that this is strictly bigger than its EnSiam counterpart (in the sense of semidefinite inequality). In conclusion, EnSiam has a smaller variance for gradients than SimSiam. It is also evident in Table \ref{Ensemble size effect_total} that this phenomenon is consistent for various numbers of augmentations $K$.

\section{Empirical Study}
\label{Empirical Study}

Here, we provide various empirical evaluations to confirm the effectiveness of EnSiam. The experiments include linear evaluation on various datasets, transfer learning for object detection, and verifying the effects of various hyperparameters.

\begin{table}
  \caption{Linear evaluation on small datasets.}
  \label{Small dataset linear evaluation}
  \centering
  \begin{tabular}{lllll}    
     Method & CIFAR-10 & CIFAR-100 & STL-10 & Tiny-ImageNet  \\
    \toprule
     Supervised & 94.22 & 74.66 & 82.55 & 59.26  \\
    \midrule
     SimCLR \cite{chen2020simple} & 84.92 & 59.28 & 85.48 & 44.38  \\
     BYOL \cite{grill2020bootstrap} & 85.82 & 57.75 & 87.45 & 42.70  \\
     MoCo v2 \cite{chen2020improved} & 86.18 & 59.51 & 85.88 & 43.36  \\
     ReSSL \cite{zheng2021ressl} & 90.20 & 63.79 & 88.25 & 46.60  \\
     SimSiam \cite{chen2021exploring} & 88.51 & 60.00 & 87.47 & 37.04  \\
     SimSiam$^\dagger$ & 90.08 & 64.08 & 89.37 & 48.95  \\
     SimSiam 4-aug & 91.01 & 66.67 & 90.40 & 49.71  \\
     \midrule
         EnSiam (4-aug) & \textbf{91.67} & \textbf{67.15} & \textbf{92.57} & \textbf{52.43}  \\
    \bottomrule
  \end{tabular}
\end{table}

\begin{table}
  \caption{Linear evaluation on ImageNet.}
  \label{ImageNet linear evaluation}
  \centering
  \begin{tabular}{llll}    
     Method & Batch Size & Epochs & Top-1 acc.  \\
    \toprule
     Supervised & 256 & 120 & 76.5  \\
    \midrule
     InstDisc \cite{wu2018unsupervised} & 256 & 200 & 58.5  \\
     LocalAgg \cite{zhuang2019local} & 128 & 200 & 58.8  \\
     MoCo v2 \cite{chen2020improved} & 256 & 200 & 67.5  \\
     MoCHi \cite{kalantidis2020hard} & 512 & 200 & 68.0 \\
     CPC v2 \cite{henaff2020data} & 512 & 200 & 63.8  \\
     PCL v2 \cite{li2020prototypical} & 256 & 200 & 67.6  \\
     AdCo \cite{hu2021adco} & 256 & 200 & 68.6  \\
     CLSA-Single \cite{wang2022contrastive} & 256 & 200 & 69.4  \\
     SimCLR \cite{chen2020simple} & 4096 & 200 & 66.8  \\
     SwAV \cite{caron2020unsupervised} & 4096 & 200 & 69.1  \\
     BYOL \cite{grill2020bootstrap} & 4096 & 200 & 70.6 \\
     WCL \cite{zheng2021weakly} & 4096 & 200 & 70.3  \\
     ReSSL \cite{zheng2021ressl} & 256 & 200 & 69.9  \\
     \midrule
     SimSiam \cite{chen2021exploring} & 256 & 200 & 70.0  \\
     SimSiam 4-aug & 256 & 200 & 69.5  \\
     \midrule
     EnSiam (4-aug) & 256 & 200 & \textbf{71.2}  \\

    \bottomrule
  \end{tabular}
\end{table}

\subsection{Baseline Settings}
\textbf{Encoder.} We used ResNet18 \cite{he2016deep} as the backbone for small datasets (CIFAR10, CIFAR100 \cite{krizhevsky2009learning}, STL10 \cite{coates2011analysis}, and Tiny-ImageNet \cite{le2015tiny}) and ResNet50 for ImageNet. When using ResNet18, we replaced the initial 7x7 stride-2 convolution with a 3x3 stride-1 convolution and removed the first max pooling operation. The channel size of the hidden layers in the projection MLP was set to 2048. For small datasets, the projection MLP was composed of two layers, while it was three layers for ImageNet. The prediction MLP consisted of two layers, and the hidden layer had a channel size of 512. Note that the settings for the projection and prediction MLPs followed those of SimSiam.

\textbf{Optimizer.} We used SGD for both pretraining and linear evaluation, and the learning rate was set in the same way as in SimSiam, i.e., $\texttt{lr} = \texttt{baselr} \times \texttt{BatchSize}/256$. Unless otherwise specified, we set the base learning rate ($\texttt{baselr}$) as 0.10. We set the momentum to 0.9 and the weight decay to $10^{-4}$. We adopted the cosine learning rate schedule and used a warm-up strategy where the warm-up period was set to 10 epochs.

\textbf{Augmentation configurations.}
The default augmentation configuration for pretraining included random cropping, random horizontal flip, color jittering with the same parameters as in \cite{chen2020improved, chen2021exploring}, random grayscale, and Gaussian blurring with $\sigma \in  [0.1, 2.0]$. The default augmentation configuration for linear evaluation was composed of random cropping and random horizontal flip.

\textbf{Misc.}
We used a batch size of 256 as default. The default training epoch of all methods was 200 for pretraining, and 100 for linear evaluation. The default $K$ was four for EnSiam and SimSiam $K$-aug. All the experiments were conducted on Nvidia RTX 3090. Most of the experiments were conducted on a single GPU, except for ImageNet (10 GPUs), STL-10 (3 GPUs), object detection (10 GPUs), and if stated otherwise. The multi-GPU training was performed using the \texttt{DistributedDataParallel} framework in PyTorch.

\subsection{Linear Evaluation Results on Small Datasets}
We applied the linear evaluation protocol, which is popular in SSL, to assess the quality of representations. Table \ref{Small dataset linear evaluation} shows the linear evaluation performance of various state-of-the-art methods. Here, the values are mostly taken from \cite{zheng2021ressl} except for SimSiam $4$-aug and EnSiam, and we have also reproduced SimSiam and reported the performance (SimSiam$^\dagger$). Note that, in this table, SimSiam $K$-aug generates the same number of augmentations as EnSiam, so we can directly see the effect of ensemble representations by comparing these two. As can be seen from the table, EnSiam achieves the highest performance on all datasets. In addition, SimSiam$^\dagger$ and SimSiam 4-aug also show high performance, but they are still inferior to the proposed method, where the gap is bigger for larger datasets. This confirms that using ensemble representations is indeed effective for the problem.

\subsection{Linear Evaluation Results on ImageNet}
We also evaluated our method on ImageNet-1k \cite{deng2009imagenet}. Again, the values are mostly taken from \cite{zheng2021ressl} except for SimSiam 4-aug and EnSiam. Table \ref{ImageNet linear evaluation} shows the top-1 accuracies of various methods. The proposed method outperforms the existing state-of-the-art (BYOL) by 0.6\% in linear evaluation, achieving the highest performance. Compared to ReSSL, which has the highest performance among existing methods with a relatively small batch size, the improvement of the proposed method is 1.2\%. SimSiam 4-aug shows similar or slightly lower performance compared to existing methods, and the performance gap between SimSiam 4-aug and EnSiam is about 1.7\%. This indicates that the proposed ensemble representations can play a significant role in representation learning, particularly for large-scale datasets.

\begin{table}
  \caption{Transfer learning for object detection (Pascal VOC).}
  \label{Transfer Learning}
  \centering
  \begin{tabular}{llll}    
     Pretraining method & AP50 & AP & AP75  \\
    \toprule
     From scratch & 60.2 & 33.7 & 33.1  \\
     Supervised pretraining & 81.3 & 53.5 & 58.8  \\
    \midrule
     SimCLR \cite{chen2020simple} & 81.8 & 55.5 & 61.4  \\
     MoCo v2 \cite{chen2020improved} & 82.3 & 57.0 & 63.3  \\
     BYOL \cite{grill2020bootstrap} & 81.4 & 55.3 & 61.1 \\
     SwAV \cite{caron2020unsupervised} & 81.5 & 55.4 & 61.4  \\
     \midrule
     SimSiam \cite{chen2021exploring} & 82.0 & 56.4 & 62.8 \\
     Simsiam (optimized) \cite{chen2021exploring} & 82.4 & 57.0 & \textbf{63.7}  \\
     \midrule
     EnSiam (4-aug) & \textbf{82.5} & \textbf{57.2} & 63.6  \\
    \bottomrule
  \end{tabular}
\end{table}

\subsection{Transfer Learning}

Table \ref{Transfer Learning} shows the performance of transfer learning for object detection. Pascal VOC \cite{everingham2010pascal} was used for this experiment, and we evaluated our method using the transfer learning code based on Detectron \cite{Detectron2018} as in MoCo \cite{chen2020improved} and SimSiam \cite{chen2021exploring}. The learning rate for fine-tuning was adjusted to find the best performance, following the practices of existing methods. Here, we mostly show the values reported in SimSiam \cite{chen2021exploring}, except for the proposed method. All methods use ResNet50 as the backbone, which is pretrained on ImageNet for 200 epochs using the respective methods. Here, EnSiam shows the highest performance in most cases. In addition, our method also shows comparable performance to SimSiam (optimized), which is a version of SimSiam that the hyperparameters are optimized for this particular task. On the other hand, the performance of EnSiam in this table is based on the same model in Table \ref{ImageNet linear evaluation}. Even so, EnSiam has higher performance in AP and AP50.

\subsection{Impacts of Hyperparameters}
\label{effects_of_hyperparameter}

In this section, we evaluate the impacts of various hyperparameters. For these experiments, we used Tiny-ImageNet. We have also evaluated SimSiam $K$-aug in this section to verify the effect of ensemble representations more closely.

\begin{table}
  \caption{Performance with various batch sizes (Tiny-ImageNet).}
  \label{batch size effect}
  \centering
  \begin{tabular}{llllllll}    
     \multicolumn{2}{c}{Method {\textbackslash} Batch size} & 8 & 16 & 32 & 64 & 128 & 256 \\
    \toprule
     \multirow{2}{*}{SimSiam $4$-aug} & Acc. (\%) & 39.78 & 44.35 & 49.51 & 51.01 & 51.19 & 49.71  \\
                                    & Time (min) & 7,543  & 4,260  & 3,019  & 2,490  & 1,165  & 1,164   \\
    \midrule
     \multirow{2}{*}{EnSiam (4-aug)} & Acc. (\%) & 47.72 & 49.32 & 51.54 & 51.65 & 51.81 & 52.43 \\
                             &Time (min) & 7,621  & 3,765  & 2,967  & 2,480  & 1,170   & 1,170 \\
    \bottomrule
  \end{tabular}
\end{table}

\textbf{Batch size.} The previous SSL studies have observed that batch size significantly impacts performance and that larger batch sizes can result in higher performance \cite{chen2020simple, grill2020bootstrap, he2020momentum, chen2021exploring}. Hence, we investigated the performance changes of EnSiam w.r.t. batch size. In Table \ref{batch size effect}, EnSiam demonstrates stable performance across a wider range of batch sizes compared to the baseline method, SimSiam $4$-aug. Even when decreasing the batch size from 256 to 8, the performance decreases only by 4.71\%, which is much more robust than SimSiam 4-aug's 9.93\% degradation. These results highlight the robustness of our method, especially in resource-constrained environments.

\textbf{Number of augmentations ($K$).} Here, we investigate the impact of $K$, which is an important hyperparameter in the proposed method. In this experiment, due to the limitation of VRAM capacity, it was not possible to perform all experiments in a single GPU with a batch size of 256. However, we observed empirically that the performance slightly decreases with multi-GPU training. To see the characteristics without the effect of multi-GPU training, we also repeated the experiments with a batch size of 64.
In Table \ref{Ensemble size effect_total}, we indicated experiments using two GPUs with `$\ast$', and those using four GPUs with `$\ast\ast$'. Here, EnSiam consistently outperforms SimSiam $K$-aug in all cases. Note that the SimSiam 2-aug is identical to the original SimSiam, and EnSiam can also work with $K=2$ where $\bar{f}$ simply becomes the average of two representations. Even in this case, the performance improvement from SimSiam is  0.47\% for a batch size of 256 and 0.6\% for that of 64.
\begin{table}
  \caption{Performance with various $K$ (Tiny-ImageNet).}
  \label{Ensemble size effect_total}
  \centering
  \begin{tabular}{llllll}
     Method {\textbackslash} $K$ & Batch size & 2 & 4 & 8 & 16  \\
    \toprule
     SimSiam $K$-aug & 256 & 48.95 & 49.71 & $49.75^{\ast}$ & $49.59^{\ast\ast}$\\
     EnSiam           & 256 & 49.42 & 52.43 & $51.97^{\ast}$ & $50.23^{\ast\ast}$\\
    \midrule
     SimSiam $K$-aug & 64 & 48.87 & 51.01 & 52.12 & 52.63\\
     EnSiam          & 64 & 49.47 & 51.65 & 52.36 & 53.23\\
    \bottomrule
  \end{tabular}
\end{table}

\begin{figure}
  \centering
  \includegraphics[width=0.68\columnwidth]{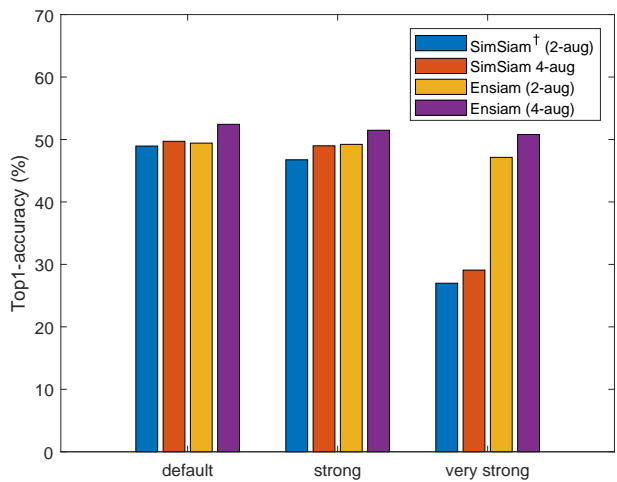}
  \caption{Performance with various augmentation settings (Tiny-ImageNet).}
  \label{Ensemble size effect for Strong Augmentation_fig}
\end{figure}

\begin{figure}
  \centering
  \includegraphics[width=\columnwidth]{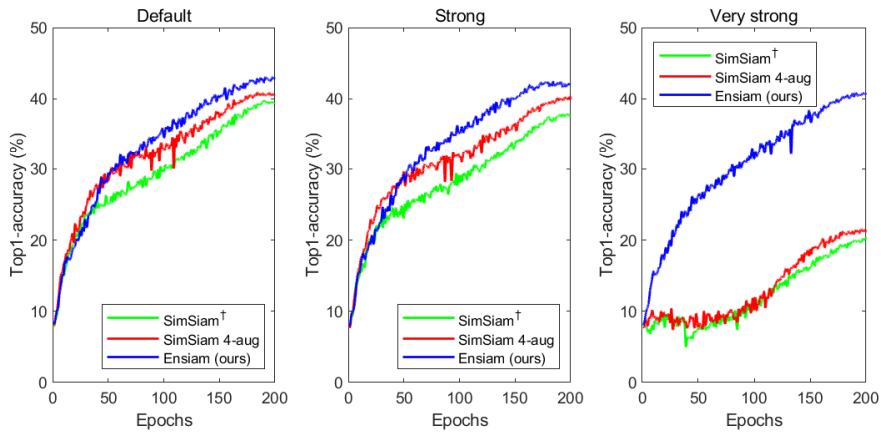}
  \caption{$K$NN accuracy trends w.r.t. training epochs under various augmentation settings (Tiny-ImageNet).}
  \label{KNN}
\end{figure}

\textbf{Augmentation settings.} In addition, we conducted experiments on various augmentation settings to evaluate the robustness of EnSiam. We designed two new sets of augmentation configurations. The first one is called the `strong' setting, in which the hyperparameters of the default augmentation configuration are tuned to generate more dynamic samples: We expanded the color jittering parameters, i.e., the range of brightness, contrast, and saturation, by 0.4, and adjusted that of Gaussian blurring to $\sigma \in  [0.2, 3.0]$. The second one is the `very strong' setting, where \texttt{RandAugment(2,5)} \cite{cubuk2020randaugment} and jigsaw with $4\times4$ grids \cite{noroozi2016unsupervised, chen2019destruction} were added to generate even more challenging samples. As shown in Figure \ref{Ensemble size effect for Strong Augmentation_fig}, EnSiam exhibits less performance degradation than SimSiam$^{\dagger}$ and SimSiam 4-aug in most cases. This demonstrates that EnSiam is more robust to augmentation settings, so we can put less effort into tuning augmentations. In Figure \ref{KNN}, we also provide the $K$-nearest neighbor ($K$NN) accuracy of each algorithm. In both the default and strong augmentation settings, EnSiam exhibits an initial slow increase in $K$NN accuracy compared to SimSiam$^{\dagger}$ and SimSiam $K$-aug. However, the accuracy of EnSiam gets higher than SimSiam$^{\dagger}$ and SimSiam $K$-aug as the training progresses. This can be attributed to the fact that EnSiam, with its ensemble representation, may change slowly during the early stages of training unlike SimSiam$^{\dagger}$ and SimSiam $4$-aug where the representations are directly handled. In the very strong setting, both SimSiam and SimSiam 4-aug struggle in the early stages of training and they still cannot catch up EnSiam in the later stages. It appears that EnSiam enables much more stable training even with the challenging augmentations.


\section{Conclusion}
We proposed a new robust SSL method, EnSiam. EnSiam generates multiple augmented views and calculates their ensemble representations to provide stable pseudo labels for learning representations. We analyzed the variance of gradients to reason the effectiveness of EnSiam from the variance-reduction perspective. Experimental results demonstrate that EnSiam achieves state-of-the-art performance in representation learning across various datasets and protocols. Furthermore, it is also shown that the proposed method can mitigate the performance degradation of SimSiam caused by changes in training configurations. EnSiam can provide higher performance more stably with less effort in hyperparameter tuning, and it can be especially useful in resource-limited environments. 
The idea of the proposed method can also be extended to other methods in self-supervised learning to improve performance, which is left as future work.

\section{Limitations}
Even though the proposed method provides more stable training characteristics and high performance, it is not without a cost. Increasing $K$ can incur higher computational and memory burdens. This can be resolved using $K=2$ or a smaller batch size, where EnSiam still beats the existing methods. Among these two, we recommend the latter approach because it gives higher performance.
Despite the improved stability in training, there is still a limitation in strong augmentation settings in that what EnSiam achieves here is more performance preservation rather than improvement. However, the proposed method is simple and flexible, and it can be combined with the methods specialized for such strong augmentations \cite{bai2022directional}. By doing so, we can expect to achieve higher performance in various conditions.

\bibliographystyle{plain}
\bibliography{myBib.bib}

\end{document}